\pdfoutput=1
\documentclass[sigconf,nonacm]{acmart}

\usepackage{graphicx}
\usepackage{amsmath}
\usepackage{booktabs}
\usepackage{multirow}
\usepackage{placeins}

\begin{document}

\title{Budgeted Task-Aware Acquisition of Dynamic Networks}

\author{Zihe Zhou}
\orcid{0009-0009-9000-7436}
\affiliation{%
  \institution{Department of Mechanical \& Industrial Engineering, University of Toronto}
  \city{Toronto}
  \state{Ontario}
  \country{Canada}
}
\email{zhchen.zhou@mail.utoronto.ca}

\begin{abstract}

Learning on dynamic graphs is difficult when changes in the underlying network are only partially observed.
Acquiring current graph information incurs observation and computational costs, making complete updates impractical under limited resources.
This paper focuses on budgeted task-aware acquisition on dynamic networks, where a model needs to decide which stale graph information to refresh for a downstream task.
We propose \textsc{Scout}, a lightweight framework that learns the task value of querying each node from the maintained graph and observation history.
Our evaluation covers one synthetic and 4 real-world dynamic networks, 2 downstream tasks, 9 acquisition baselines, and several query budgets.
\textsc{Scout} achieves the highest mean downstream performance in 15 of the 21 benchmark--budget settings and ranks within the top two in 19 of them.
Its average rank is 1.4, compared with 3.2 for the strongest baseline.
On the same dynamic network, task-matched acquisition improves link-prediction AUC by 0.005--0.007 and node-classification accuracy by 0.007--0.017 over task-mismatched acquisition.
These results show that useful graph observations depend on the downstream task and that limited observation budgets can be allocated effectively by learning directly from downstream utility.

\end{abstract}

\keywords{Dynamic graphs, Partially observed graphs, Active information acquisition, Budgeted graph querying
}

\maketitle

\section{Introduction}
\label{sec:introduction}

Graphs provide a natural way to represent interactions among entities. 
In many Web-based systems, including online social networks~\cite{kossinets2006empirical} and communication networks~\cite{holme2012temporal}, graphs evolve over time as nodes and edges are added, removed, or changed~\cite{zheng2025survey}.
In practice, however, changes in the underlying system may not be immediately reflected in the graph available for downstream tasks~\cite{ghalebi2018dynamic}.
Acquiring up-to-date graph information incurs additional costs.
Under a limited budget, continuously observing every node in a large Web network is impractical.
The key challenge is therefore to decide which nodes to query for updated information.
This decision should account for how the acquired information benefits the downstream task.

Several lines of research are relevant to budgeted information acquisition in dynamic graphs.
Budgeted observation has been considered in dynamic networks for specific objectives, including influence maximization~\cite{zhuang2013influence} and network representation learning~\cite{han2019network}. 
Other studies reconstruct missing node and edge information from partial observations~\cite{ghalebi2018dynamic,wang2023networked}.
Active acquisition methods select costly node states or features to improve a downstream prediction task~\cite{sterchi2023active,guney2025active}. 
Web crawling studies also optimize data freshness under limited resources~\cite{cho2003effective}.
However, it remains difficult to estimate the downstream utility of node queries in partially observed dynamic graphs.
In evolving Web networks, querying a node may reveal changes in its state, incident edges, and previously unknown neighbors.
These updates may also affect downstream predictions through graph dependencies.
Thus, node queries may provide different benefits to the downstream task.
We therefore study task-aware node acquisition under a limited budget in dynamic graphs.
Our main contributions are:

1. We introduce budgeted task-aware acquisition for partially observed dynamic networks. The value of refreshing graph information is defined by its expected utility for a downstream task, rather than by structural change alone.

2. We compare downstream task utility with graph change as supervision signals through controlled experiments. Downstream task utility often provides a more effective supervision signal for learning acquisition decisions.

3. We develop a lightweight learned scorer to estimate query utility. The proposed method is evaluated on synthetic and real-world dynamic networks under multiple observation budgets.

\section{Methodology}
\label{sec:methodology}

\subsection{Problem Framework} 
\label{sec:problem_framework}

Consider a dynamic graph $G_t = (V_t, E_t, X_t)$ at time $t$, where $V_t$ is the set of nodes, $E_t$ is the set of edges and $X_t=\{x_v^t:v\in V_t\}$ contains the node states.
The downstream task does not directly observe the complete graph $G_t$.
Instead, it maintains a potentially stale graph $\widehat{G}_t = (\widehat{V}_t, \widehat{E}_t, \widehat{X}_t)$, where $\widehat X_t = \{\widehat x_v^{\, t}: v \in \widehat V_t\}$ contains the most recently observed node states.
We assume that the initial graph is fully observed, such that $\widehat G_0 = G_0$.
For each known node $v \in \widehat{V}_t$, let $\tau_v^t \leq t$ denote its most recent observation time.
If $v$ has been previously queried, then $\tau_v^t \geq 0$ and $\widehat x_v^{\, t} = x_v^{\tau_v^t}$.

At time $t$, a subset of known nodes $Q_t \subseteq \widehat{V}_t$ is queried.
Querying $v \in Q_t$ reveals an observation $\mathcal O_t(v)=\bigl(a_v^t,x_v^t,N_v^t\bigr)$ from the graph $G_t$, where $a_v^t \in \{0, 1\}$ indicates whether $v$ currently exists, $x_v^t$ is its current state, and $N_v^t$ is its current neighbors.
If $a_v^t = 0$, the observation indicates that $v$ has been removed from the graph $G_t$.
The node and its incident edges are then removed from the graph $\widehat G_t$.
Let $\widehat N_v^{\, t}$ denote the neighbors of $v$ in $\widehat G_t$.
If $a_v^t = 1$, comparing $N_v^t$ with the stored neighborhood $\widehat N_v^t$ identifies the added neighbors $\Delta N_v^{+}=N_v^t\setminus\widehat N_v^t$ and the removed neighbors $\Delta N_v^{-}=\widehat N_v^t\setminus N_v^t$ in $G_t$.
Previously unknown nodes revealed by $\Delta N_v^{+}$ are added to the maintained graph.
For each newly discovered node $u \in \Delta N_v^{+}$, we initialize $\tau_u^t = -1$.
Its node state $x_u^t$ and full neighborhood $N_u^t$ remain unobserved, and it can be queried at a later time step.

We measure the value of a query set $Q_t$ by its effect on the downstream task.
A query is useful not merely when it reveals a graph change, but when the revealed information improves the downstream prediction. 
We therefore define query utility directly through downstream loss.
Let $\widehat G_t^{\, Q_t}$ denote the maintained graph after incorporating the observations revealed by querying $Q_t$.
In this work, we assume that the updated maintained graph is carried forward to the next time step.
Thus, $\widehat G_{t+1}=\widehat G_t^{\,Q_t}$ before queries are made at time $t+1$.
Let $f$ denote a downstream prediction model, $y_t$ the task labels at time $t$, and $\mathcal{L}$ the prediction loss.
We define the task utility of $Q_t$ as the expected reduction in downstream loss:
\begin{equation}
\label{eq:task_utility}
U_t(Q_t)
=
\mathbb{E} \left[ \mathcal L(f(\widehat G_t), y_t) - \mathcal L(f(\widehat G_t^{\, Q_t}), y_t) \mid \mathcal H_t \right],
\end{equation}
where $\mathcal H_t$ contains the maintained graph $\widehat G_t$ and all observations available before $Q_t$ is selected.
The budgeted acquisition problem is therefore
\begin{equation}
\begin{aligned}
Q_t^*
&=
\arg\max_{Q_t \subseteq \widehat V_t} U_t(Q_t) \\
\text{s.t.}\quad
& |Q_t| = \min(B_t, |\widehat V_t|).
\end{aligned}
\end{equation}
We assume a unit cost for each node query, such that exactly $b_t = \min(B_t, |\widehat V_t|)$ nodes are queried at time $t$.
Importantly, the current state of a node is unknown before it is queried. 
Therefore, $U_t(Q_t)$ cannot be evaluated directly at decision time and must be estimated from the stale graph and previous observation history.

\subsection{\textsc{Scout}: Learning Task-Aware Acquisition} 
\label{sec:learning_task_aware_acquisition}

\begin{figure*}[t]
    \centering
    \includegraphics[width=\textwidth]{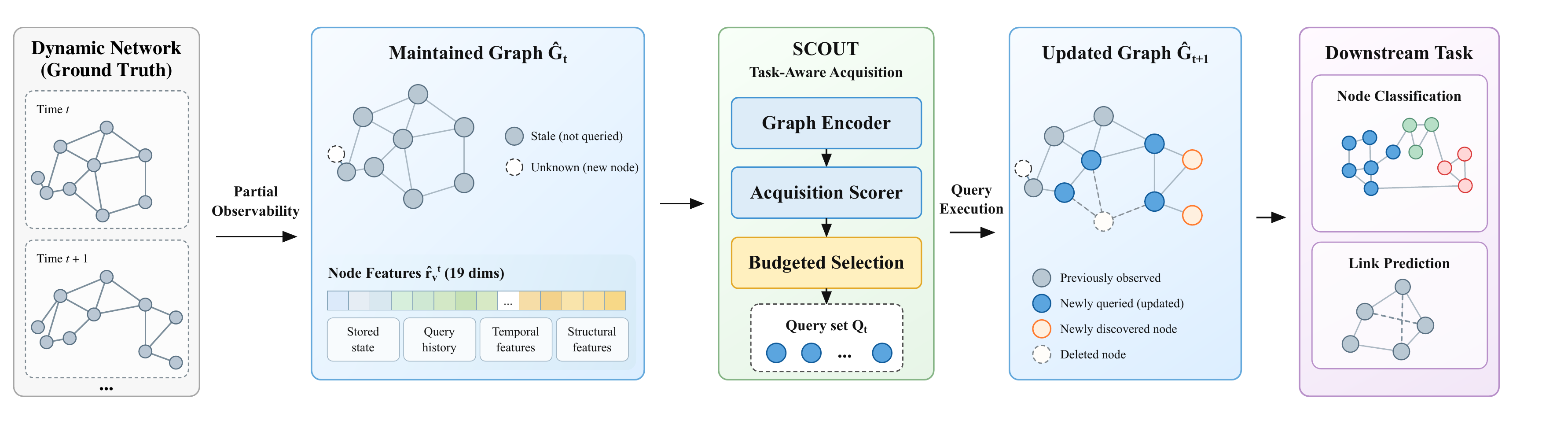}
    \caption{Overview of \textsc{Scout}, which learns task-aware acquisition scores to query and update a partially observed dynamic graph under a limited budget.}
    \label{fig:scout_overview}
\end{figure*}

To solve this acquisition problem, we propose \textsc{Scout}, a task-aware acquisition framework for dynamic graphs.
\textsc{Scout} is applicable to downstream tasks with a well-defined evaluation objective.
It consists of five stages: maintained feature construction, maintained-graph encoder, task-utility learning, budgeted query selection, and query execution with graph update.
Figure~\ref{fig:scout_overview} illustrates the overall workflow of \textsc{Scout}.

\subsubsection{Maintained Feature Construction}
\label{sec:maintained_feature_construction}

For each node $v \in \widehat V_t$, \textsc{Scout} constructs a feature vector $\widehat r_v^{\, t} \in \mathbb{R}^{19}$ using the information available before querying $v$. 
This vector summarizes the stored node state, query history, temporal uncertainty, maintained neighbors, and task-model signals. 
Representative features in $\widehat r_v^{\, t}$ include transformed measures of node staleness, query frequency, and maintained degree.
We collect the feature vectors of all known nodes and standardize them into a matrix $\widehat R_t \in \mathbb{R}^{|\widehat V_t|\times 19}$ with respect to the training data.

\subsubsection{Maintained-Graph Encoder}
\label{sec:maintained_graph_encoder}

\textsc{Scout} uses $\widehat R_t$ and the structure of $\widehat G_t$ to learn graph-aware node representations.
An input MLP first maps these features to hidden representations, where $H_t^{(0)}=\phi_{\mathrm{in}}(\widehat R_t) \in \mathbb{R}^{|\widehat V_t| \times d}$.
Here $\phi_{\mathrm{in}}$ is the input MLP shared across all nodes.

Let $\widehat A_t \in \{0, 1\}^{|\widehat V_t| \times |\widehat V_t|}$ denote the adjacency matrix of $\widehat G_t$.
$\widehat A_t(u, v) = 1$ if nodes $u$ and $v$ are connected in $\widehat G_t$, 0 otherwise.
Let \(\widehat d_v^{\,t}=\sum_{u\in\widehat V_t}\widehat A_t(u,v)\) denote the degree of node \(v\) in the maintained graph.
The diagonal degree matrix of \(\widehat A_t+I\) is \(\widetilde D_t\), where \(\widetilde D_t(v,v)=1+\widehat d_v^{\,t}\).
We define the normalized maintained adjacency of $\widehat G_t$ as $\widehat A_t^{\, \ast} = \widetilde D_t^{-1/2} (\widehat A_t + I) \widetilde D_t^{-1/2}$.
\textsc{Scout} then applies two graph-convolution layers with skip connections.
For layer $\ell \in \{0, 1\}$, 
\begin{equation}
H_t^{(\ell + 1)} = H_t^{(\ell)} + \mathrm{ReLU}(H_t^{(\ell)}W_{\mathrm{self}}^{(\ell)} + \widehat A_t^{\, \ast} H_t^{(\ell)} W_{\mathrm{neigh}}^{(\ell)})
\end{equation}
Here, $W_{\mathrm{self}}^{(\ell)}$ applies a direct transformation to each node representation, while $W_{\mathrm{neigh}}^{(\ell)}$ transforms the normalized aggregation over the maintained graph. 
Since $\widehat A_t^{\,\ast}$ includes self-loops, the aggregation also contains the node itself with weight $1/(1+\widehat d_v^{\,t})$.
The skip connection adds the previous node representation to the graph-convolution output.
Let $h_v^t$ denote the representation of node $v$ in $H_t^{(2)}$. 
A shared scoring MLP maps each node representation to a scalar acquisition score $s_v^t = \phi_{\mathrm{score}}(h_v^t), v \in \widehat V_t$.
The score $s_v^t$ estimates the relative task value of querying node $v$ using only known information.

\subsubsection{Task-Utility Learning}
\label{sec:task_utility_learning}

During training, \textsc{Scout} constructs node-level utility targets for a sampled set of candidate nodes $\mathcal{C}_t$.
For each candidate $v \in \mathcal{C}_t$, we form a singleton counterfactual graph by applying only its current observation to the maintained graph, where $\widehat G_t^{\{v\}} = \widehat G_t \oplus \mathcal{O}_t(v)$. 
These counterfactual observations are used only to construct training targets and are unavailable to \textsc{Scout} at inference time.
The corresponding singleton utility is $u_v^t = \mathcal{L}(f(\widehat G_t), y_t) - \mathcal{L}(f(\widehat G_t^{\{v\}}), y_t)$.
A positive value of $u_v^t$ indicates that querying node $v$ reduces the downstream prediction loss.
This provides a node-level training target for the expected task utility defined in Eq.\eqref{eq:task_utility}.
\textsc{Scout} then standardizes these utilities, $\tilde{u}_v^t = (u_v^t - \mu_t)/(\max(\sigma_t, \epsilon))$.
$\mu_t$ and $\sigma_t$ are the mean and standard deviation over the sampled candidate set $\mathcal{C}_t$.
\textsc{Scout} is then trained with a regression term on the standardized utilities together with a ranking term over $\mathcal{C}_t$:

\begin{equation}
\label{eq:scout_loss}
\mathcal{L}_{\mathrm{Scout}}
=
\frac{1}{|\mathcal{C}_t|}
\sum_{v \in \mathcal{C}_t}
\big(s_v^t-\tilde{u}_v^t\big)^2
+
\lambda_{\mathrm{rank}} D_{\mathrm{KL}}(p^t \| q^t).
\end{equation}
where $p^t=\mathrm{softmax}(\tilde{u}^t/\tau)$ and $q^t=\mathrm{softmax}(s^t)$ are the target and predicted ranking distributions over $\mathcal{C}_t$, respectively.
In addition, the hyperparameter $\lambda_{\mathrm{rank}} \geq 0$ controls the contribution of the ranking term to the training objective.

\subsubsection{Budgeted Query Selection}
\label{sec:budgeted_query_selection}

Given the acquisition score $s_v^t$, \textsc{Scout} selects a query set $Q_t$ under the budget $B_t$. 
Relying only on the learned score may repeatedly favor previously observed nodes and leave other nodes unexplored.
To encourage occasional exploration, we add a small bonus on node staleness, $\delta_v^t = t-\tau_v^t$.
The adjusted selection score is $\widetilde s_v^{t} = s_v^t + \lambda \log(1 + \delta_v^t)$, 
where $\lambda$ controls the strength of exploration.
We set $\lambda = 0.05$ as default.

\textsc{Scout} ranks all known nodes according to the adjusted score $\widetilde s_v^t$.
Let $b_t = \min(B_t, |\widehat V_t|)$ denote the number of nodes that can be queried at time $t$.
\textsc{Scout} selects the $b_t$ nodes with the largest adjusted scores, $Q_t = \mathrm{Top}_{b_t}\{\widetilde s_v^t: v \in \widehat V_t\}$.

\subsubsection{Query Execution with Graph Update}
\label{sec:query_execution_with_graph_update}

After selecting $Q_t$, \textsc{Scout} queries each selected node from the current graph $G_t$.
The returned observations are incorporated into the maintained graph according to the update rules defined in Section~\ref{sec:problem_framework}, yielding the updated graph $\widehat G_t^{\, Q_t}$.
The updated graph is carried forward to the next time step, such that $\widehat G_{t+1} = \widehat G_t^{\, Q_t}$.

\section{Experiments}
\label{sec:experimental}

\subsection{Experimental Setup}
\label{sec:experiemental_setp}

\textbf{Datasets.}
We construct one synthetic dynamic graph, \texttt{drift\_sbm} and use four real-world dynamic graph datasets from SNAP for evaluation.
\texttt{drift\_sbm} contains 400 nodes, four communities, and 20 time steps, with node communities and edges changing over time.
We use this benchmark for node classification and report accuracy.
The four SNAP datasets are AS-733 \cite{leskovec2005graphs}, Reddit \cite{kumar2018community}, email-Eu-core-temporal \cite{paranjape2017motifs}, and MathOverflow \cite{paranjape2017motifs}.
Link prediction on four real-world datasets is evaluated by AUC.
\textbf{Baselines.}
We compare \textsc{Scout} with six heuristic acquisition methods based on random selection, degree, staleness, degree--staleness, historical change rate, and predictive uncertainty.
We also include round-robin, closeness-centrality, and CPNE-style probing following prior work on partial monitoring of dynamic networks \cite{han2019network}.

\subsection{Budgeted Acquisition Performance}
\label{sec:budgeted_acquisition_performance}

\begin{figure*}[t]
    \centering
    \includegraphics[width=\textwidth]{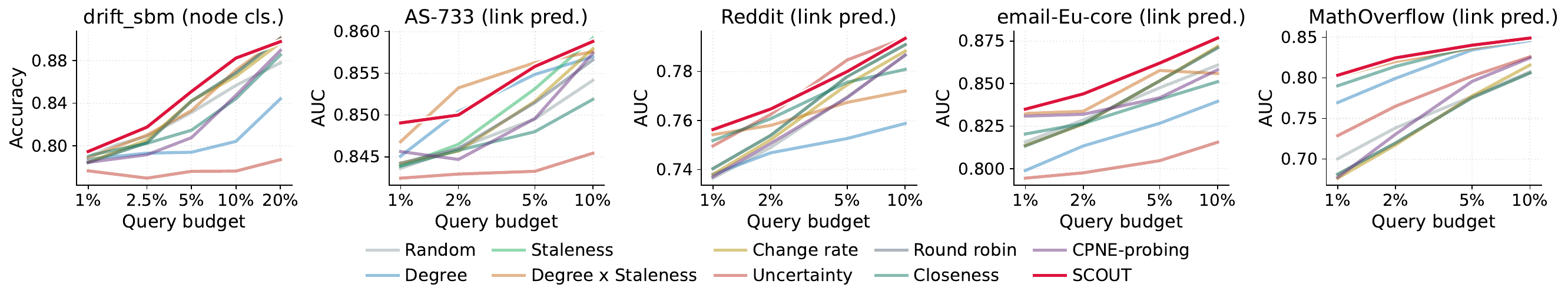}
    \caption{Downstream performance of 10 acquisition methods across query budgets.
            Budgets denote the percentage of nodes queried at each time step.
            The left panel reports node-classification accuracy on \texttt{drift\_sbm}; 
            the four right panels report link-prediction AUC on real-world networks.}
    \label{fig:baseline_curves}
\end{figure*}

Figure~\ref{fig:baseline_curves} compares the downstream performance of 10 acquisition methods across query budgets.
Performance generally improves as more nodes are queried.
\textsc{Scout} achieves the highest mean performance in 15 of the 21 benchmark--budget settings.
It ranks first or second in 19 settings and its average rank is 1.4.
On \texttt{drift\_sbm}, \textsc{Scout} reaches an accuracy of 0.882 at the 10\% budget, compared with 0.871 for the strongest baseline.
On Email-Eu-core, \textsc{Scout} improves over the strongest baseline by 0.01 AUC at the 2\% budget.

The gains are relatively smaller on MathOverflow, AS-733 and Reddit.
\textsc{Scout} remains the strongest method on MathOverflow across all budgets.
However, its performance margins over the baselines are modest.
On AS-733, \textsc{Scout} is within 0.004 AUC of the strongest baseline at all budgets.
On Reddit, \textsc{Scout} is competitive at 1\%, 2\%, and 10\%, but uncertainty-based acquisition performs better at 5\% (0.785 versus 0.780 AUC).
These results show that task-aware acquisition is consistently among the strongest methods across dynamic networks. 
However, its advantage becomes smaller when strong heuristics already identify informative nodes.

\subsection{Ablation Study}
\label{sec:ablation_study}

Table~\ref{tab:ablation} ablates two components of \textsc{Scout}: the supervision target and the model capacity of the acquisition scorer.
For supervision ablation, we replace the downstream task utility with a structural graph change while keeping the model architecture and evaluation protocol unchanged.
For capacity ablation, we compare \textsc{Scout} with 39-parameter and 321-parameter acquisition scorers.
Each entry reports the paired mean difference $\Delta=\textsc{Scout}-\text{variant}$, so positive values favor \textsc{Scout}.
Task-utility supervision performs better than change-based supervision in 10 of the 16 real-world benchmark--budget settings.
The differences are positive at every budget on MathOverflow.
On MathOverflow, task-utility supervision improves AUC by 0.0068 and 0.0050 at the 1\% and 2\% budgets, respectively.
On Email-Eu-core, the largest gain is 0.0130 at 1\%, while the two higher budgets show small negative differences.
Change-based supervision is better on Reddit at 10\% (by 0.0118 AUC) and on AS-733 at 2\% (by 0.0027 AUC).
These results show that task-utility supervision is most helpful at low budgets.
However, its benefits vary across networks.
The results of capacity ablation show no consistent relationship with performance.
The 321-parameter scorer is substantially worse on Reddit at the 5\% and 10\% budgets, where \textsc{Scout} improves AUC by 0.0136 and 0.0248.
In contrast, the 39-parameter scorer performs better on AS-733 at all four budgets.
These mixed results indicate that larger scorers are not uniformly better.

\begin{table}[t]
\centering
\footnotesize
\renewcommand{\arraystretch}{1.0}
\caption{Ablation results for the supervision target and acquisition scorer capacity under different query budgets.
Values are paired mean differences $\Delta=\textsc{Scout}-\text{variant}$, where positive values favor the default \textsc{Scout}.
Bold marks these positive values.
For the supervision ablation, the variant uses structural-change supervision.
For the capacity ablation, the variants use 39-parameter and 321-parameter scorers.
The 2/2.5\% column denotes 2\% on real networks and 2.5\% on \texttt{drift\_sbm}.}
\label{tab:ablation}

\begin{tabular*}{\columnwidth}{@{\extracolsep{\fill}}lcccc@{}}
\toprule
Dataset / variant & 1\% & 2/2.5\% & 5\% & 10\% \\
\midrule

\multicolumn{5}{@{}l}{\textit{Supervision: structural-change target}} \\
AS-733
& $-0.0006$ & $-0.0027$ & $\mathbf{+0.0017}$ & $\mathbf{+0.0029}$ \\
Reddit
& $\mathbf{+0.0063}$ & $\mathbf{+0.0023}$ & $-0.0002$ & $-0.0118$ \\
Email-Eu
& $\mathbf{+0.0130}$ & $\mathbf{+0.0043}$ & $-0.0021$ & $-0.0027$ \\
MathOverflow
& $\mathbf{+0.0068}$ & $\mathbf{+0.0050}$ & $\mathbf{+0.0011}$ & $\mathbf{+0.0012}$ \\

\midrule
\multicolumn{5}{@{}l}{\textit{Scorer capacity}} \\
\texttt{drift\_sbm} (321 params)
& $\mathbf{+0.0009}$ & $-0.0004$ & $-0.0055$ & $\mathbf{+0.0034}$ \\
\texttt{drift\_sbm} (39 params)
& $\mathbf{+0.0032}$ & $-0.0008$ & $-0.0010$ & $-0.0107$ \\
AS-733 (321 params)
& $-0.0016$ & $\mathbf{+0.0003}$ & $-0.0015$ & $\mathbf{+0.0002}$ \\
AS-733 (39 params)
& $-0.0046$ & $-0.0072$ & $-0.0037$ & $-0.0023$ \\
Reddit (321 params)
& $\mathbf{+0.0037}$ & $\mathbf{+0.0039}$ & $\mathbf{+0.0136}$ & $\mathbf{+0.0248}$ \\
Reddit (39 params)
& $-0.0059$ & $-0.0001$ & $\mathbf{+0.0070}$ & $\mathbf{+0.0058}$ \\

\bottomrule
\end{tabular*}
\end{table}

\subsection{Task Dependence of Acquisition}
\label{sec:task_dependence}

We examine the relationship between the acquisition value and the downstream task on AS-733 using link prediction and node classification.
For each downstream task, we compare \textsc{Scout} trained on the same task with \textsc{Scout} trained on the other task.
The acquisition architecture and evaluation protocol of both \textsc{Scout} are unchanged.
Task-matched training outperforms task-mismatched transfer at every budget.
It wins on all five test windows in 7 of the 8 task--budget settings.
For link prediction, task matching improves AUC by 0.005--0.007 across the 1--10\% budgets.
For node classification, the improvement ranges from 0.007 to 0.017 in accuracy.
The mismatched method also performs below the degree--staleness baseline at every budget.
These results show that useful graph observations are task dependent and support learning acquisition scores directly from downstream task utility.

\section{Discussion and Conclusion}
\label{sec:discussion_conclusion}

Prior work on Web crawling and network monitoring has used data freshness and graph changes to guide limited observations~\cite{cho2003effective,han2019network}.
Other studies have considered budgeted acquisition for specific downstream tasks~\cite{zhuang2013influence,sterchi2023active}.
These studies support selective observation under limited budgets, but use different acquisition objectives.
However, a general framework for task-aware acquisition across different downstream prediction tasks remains lacking.
\textsc{Scout} extends these approaches by defining acquisition value through downstream prediction loss across different tasks.
Our supervision ablation shows that downstream task utility often provides better acquisition performance than structural change.
This suggests that structural change alone may not be sufficient to identify the most valuable queries for downstream prediction.
The difference is important at low budgets, where each query must be selected carefully.
By learning downstream task utility, \textsc{Scout} can better prioritize observations under limited budgets.

Several limitations remain.
\textsc{Scout} learns from singleton counterfactual utilities. 
These utilities approximate the value of individual queries but do not model interactions among multiple queried nodes.
The formulation also assumes unit query costs, an initially observed graph, and a task-specific acquisition model.
Future work may extend this framework to heterogeneous acquisition costs, set-level utility learning, and transfer across downstream tasks.

In conclusion, this work formulates budgeted task-aware acquisition for partially observed dynamic networks.
We introduce \textsc{Scout}, a lightweight framework that learns the value of node queries from downstream task utility.
Experiments on synthetic and real-world dynamic networks show that \textsc{Scout} performs competitively across various query budgets.
Ablation results show that downstream task utility often provides a more effective acquisition target than structural change, particularly at low budgets.
Together, these findings support the importance of aligning graph acquisition with downstream objectives.

\bibliographystyle{ACM-Reference-Format}
\bibliography{refs}

@article{holme2012temporal,
  title={Temporal networks},
  author={Holme, Petter and Saram{\"a}ki, Jari},
  journal={Physics reports},
  year={2012},
  publisher={Elsevier}
}

@article{ghalebi2018dynamic,
  title={Dynamic network model from partial observations},
  author={Ghalebi, Elahe and Mirzasoleiman, Baharan and Grosu, Radu and Leskovec, Jure},
  journal={Advances in Neural Information Processing Systems},
  year={2018}
}

@article{zheng2025survey,
  title={A survey of dynamic graph neural networks},
  author={Zheng, Yanping and Yi, Lu and Wei, Zhewei},
  journal={Frontiers of Computer Science},
  year={2025},
  publisher={Springer}
}

@inproceedings{zhuang2013influence,
  title={Influence maximization in dynamic social networks},
  author={Zhuang, Honglei and Sun, Yihan and Tang, Jie and Zhang, Jialin and Sun, Xiaoming},
  booktitle={2013 IEEE 13th international conference on data mining},
  year={2013},
  organization={IEEE}
}

@inproceedings{han2019network,
  title={Network Embedding under Partial Monitoring for Evolving Networks},
  author={Han, Yu and Tang, Jie and Chen, Qian},
  booktitle={IJCAI},
  year={2019}
}

@inproceedings{wang2023networked,
  title={Networked time series imputation via position-aware graph enhanced variational autoencoders},
  author={Wang, Dingsu and Yan, Yuchen and Qiu, Ruizhong and Zhu, Yada and Guan, Kaiyu and Margenot, Andrew and Tong, Hanghang},
  booktitle={Proceedings of the 29th ACM SIGKDD Conference on Knowledge Discovery and Data Mining},
  year={2023}
}

@article{sterchi2023active,
  title={Active querying approach to epidemic source detection on contact networks},
  author={Sterchi, Martin and Hilfiker, Lorenz and Gr{\"u}tter, Rolf and Bernstein, Abraham},
  journal={Scientific Reports},
  year={2023},
  publisher={Nature Publishing Group UK London}
}

@article{guney2025active,
  title={Active feature acquisition via explainability-driven ranking},
  author={Guney, Osman Berke and Saichandran, Ketan Suhaas and Elzokm, Karim and Zhang, Ziming and Kolachalama, Vijaya B},
  journal={Proceedings of machine learning research},
  year={2025}
}

@article{cho2003effective,
  title={Effective page refresh policies for web crawlers},
  author={Cho, Junghoo and Garcia-Molina, Hector},
  journal={ACM Transactions on Database Systems (TODS)},
  year={2003},
  publisher={ACM New York, NY, USA}
}

@inproceedings{kumar2018community,
  title={Community interaction and conflict on the web},
  author={Kumar, Srijan and Hamilton, William L and Leskovec, Jure and Jurafsky, Dan},
  booktitle={Proceedings of the 2018 World Wide Web Conference on World Wide Web},
  year={2018},
  organization={International World Wide Web Conferences Steering Committee}
}

@inproceedings{leskovec2005graphs,
  title={Graphs over time: densification laws, shrinking diameters and possible explanations},
  author={Leskovec, Jure and Kleinberg, Jon and Faloutsos, Christos},
  booktitle={Proceedings of the eleventh ACM SIGKDD international conference on Knowledge discovery in data mining},
  year={2005}
}

@inproceedings{paranjape2017motifs,
  title     = {Motifs in Temporal Networks},
  author    = {Paranjape, Ashwin and Benson, Austin R. and Leskovec, Jure},
  booktitle = {Proceedings of the Tenth ACM International Conference on Web Search and Data Mining (WSDM)},
  year      = {2017},
}

@article{kossinets2006empirical,
  title={Empirical analysis of an evolving social network},
  author={Kossinets, Gueorgi and Watts, Duncan J},
  journal={science},
  year={2006},
  publisher={American Association for the Advancement of Science}
}

\end{document}